# Backswimmer Inspired Miniature Robot with Buoyancy Auto-Regulation through Controlled Nucleation and Release of Microbubbles

*Dror Kobo, Bat-El Pinchasik\**

D. Kobo, B. E. Pinchasik
Tel-Aviv University
School of Mechanical Engineering
Faculty of Engineering
Tel-Aviv, Israel

\* Correspondence: pinchasik@tauex.tau.ac.il

**Keywords**: Buoyancy regulation, bubbles, 3D printing, robotic swimmers, controlled nucleation, electrolysis

## Abstract

The backswimmer fly is an aquatic insect, capable of regulating its buoyancy underwater. Its abdomen is covered with hemoglobin cells, used to bind and release oxygen, reversibly. Upon entering water, the fly entraps an air bubble in a superhydrophobic hairy structure on its abdomen for respiration. This bubble, however, can change its volume through regulated oxygen flow from the abdominal hemoglobin cells. In this way, it can reach neutral buoyancy without further energy consumption. In this study, we develop a small, centimeter scale, backswimmer inspired robot (BackBot) with auto-buoyancy regulation through controlled nucleation and release of microbubbles. The bubbles nucleate and grow directly on onboard electrodes through electrolysis, regulated by low voltage. We use 3D printing to introduce a three-dimensional bubble-entrapping cellular structure, in order to create a stable external gas reservoir. To reduce buoyancy forces, the bubbles are released through linear mechanical vibrations, decoupled from the robot's body. Through pressure sensing and a Proportional Integral Derivative control loop mechanism, the robot auto-regulates its buoyancy to reach neutral floatation underwater within seconds. This mechanism can promote the replacement of traditional and physically larger buoyancy regulation systems, such as pistons and pressurized tanks, and to enable the miniaturization of Autonomous Underwater Vehicles.

## Introduction

In nature, one can find a variety of materials and mechanical systems that are evolutionarily optimized for underwater locomotion and for the regulation of buoyancy.[1] Most well-known are fish, sharks and other aquatic species relying on gas-filled swim bladders and fine tuning of chemical composition in internal organs, respectively.[2,3] However, a surprising mechanism for regulating buoyancy is demonstrated by the aquatic insect *Notonectidae, Anisops*, commonly known as the backswimmer.[4] By regulating the amount of oxygen bound to hemoglobin found in tracheae cells in its body, the backswimmer alters the volume of an external gas bubble, trapped in a superhydrophobic hairy structure on its abdomen (**Figure 1**).

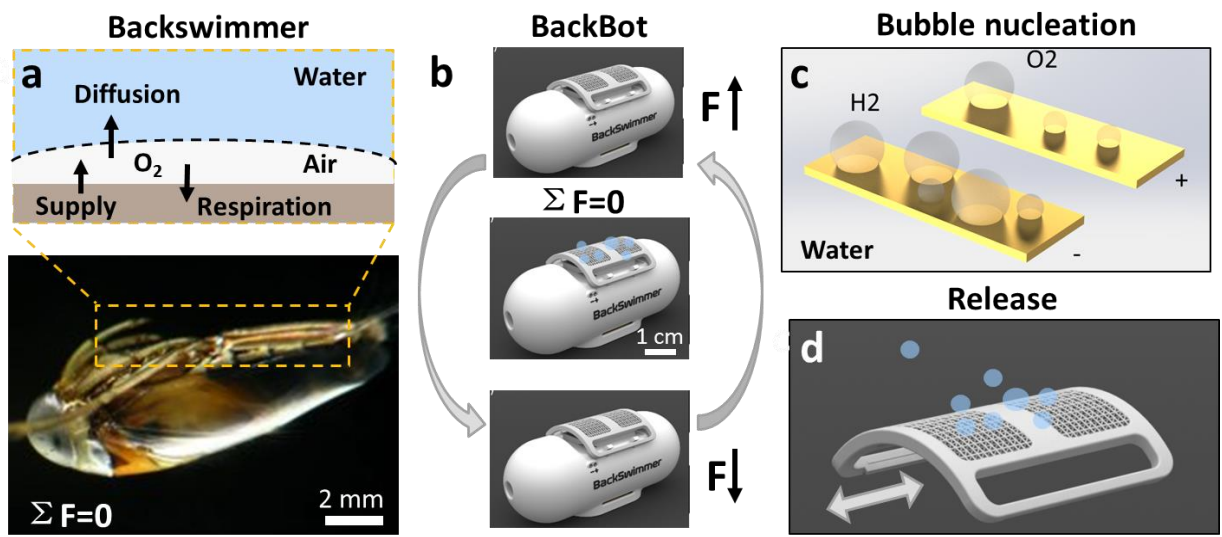

**Figure 1.** a) The aquatic insect backswimmer (Notonectidae, Anisops) can regulate its buoyancy underwater by adjusting the volume of a bubble attached to its abdomen. b) A backswimmer-inspired robot can regulate its buoyancy in a reversible way, to achieve neutral floatation underwater. c) To increase buoyancy forces, microbubbles nucleate and grow via splitting of water into oxygen and hydrogen through electrolysis. d) To reduce buoyancy, bubbles are released from a canopy cellular surface through mechanical linear vibrations. Image of the backswimmer fly is reproduced with permission.[4]

The bubble is stabilized by the superhydrophobic hairs, and is primarily used for respiration.[5,6] Therefore, it was a surprising finding that this bubble is also used to regulate buoyancy and even allows the backswimmer to reach neutral buoyancy underwater. Insects are usually negatively buoyant, meaning their density is slightly higher than water.[7] Therefore, it is possible for them to reverse the buoyancy forces underwater by small variations in the air plastron attached to their body.

Locomotion using bubbles in synthetic systems has been previously demonstrated from the nano to macro scales.[8–10] Many of these studies used bubble nucleation through catalysis.[11]

For example, the splitting of $H_2O_2$ on metals [12] or alloys [13] is convenient for propelling small colloids and particles. Urea and glucose have been demonstrated as biofuels, and enzymes as catalysts, to propel particles in a more environmentally friendly manner.[14] In order to eliminate the dependence on fuel, which may pollute and eventually deplete, other mechanisms to induce bubble nucleation are of interest. For this, electrolysis is an attractive choice, relying on the splitting of water into oxygen and hydrogen in the form of bubbles.[15] While many of these studies focused on particle propulsion due to bubble generation, there are only few examples for production of bubbles for regulating buoyancy.[16] While bubble attachment to particles is broadly used in order to induce floatation in separation processes,[17] the use of bubbles to regulate buoyancy in robotic systems, remained unexplored. Nevertheless, bubbles can generate a broad range of buoyancy forces, depending on their size, down to nN scale for sub-millimeter bubbles. Therefore, they withhold great potential to promote the miniaturization of robotic devices, offering fine-tuning of buoyancy.

Bubbles' adhesion to surfaces increases with the hydrophobicity of the surface,[18,19] density of surface cavities,[20] roughness,[21] and local decrease in pressure.[22,23] On the contrary, detaching gas bubbles from submerged surfaces can be achieved by electrowetting,[24] mechanical forces,[25,26] and when buoyancy forces overcome pinning forces.[27] In addition, when bubbles go through rapid inertial growth, they may detach early from the surface.[28] Otherwise, surface-anchored bubbles in steady state underwater change volume over time due to diffusion of gas into and from the bulk water.[29] This, for example, limits the time some insects and other arthropods spend underwater, relying on an air plastron for breathing.[30]

Since insects are generally small, up to few centimeters, and light, with up to few grams in weight, they are subjected to interfacial forces, especially at the water-air-solid interface.[31–35] While these forces can be deadly for small insects, e.g. when flying insects get trapped at the water-air interface, many insects adapted to these forces in terms of materials, design and modes of locomotion.[35–38] Such adaptation mechanisms in nature can inspire and promote the development and miniaturization of small robots, subjected to the same force.[39]

In this study, we design, fabricate and develop a small, centimeter scale, robot that can auto-regulate its buoyancy and achieve neutral floatation underwater (Figure 1b). Bubble nucleation and growth is driven by the splitting of water into oxygen and hydrogen on onboard electrodes (Figure 1c). The bubbles are entrapped and stabilized on a three-dimensional cellular structure at the top of the robot. The release of bubbles is achieved by mechanical vibrations (Figure 1d). An ad-hoc device is designed to decouple the mechanical vibrations from the robot's body, to eliminate its possible destabilization and rollover. The mechanism for controlling the detachment of the bubbles relies on linear oscillations. Overall, the BackBot is capable of self-regulating its buoyancy underwater, by controlled nucleation, growth and detachment of gas bubbles. This principle can be used to miniaturize autonomous underwater vehicles with self-regulation of buoyancy, and to develop small swimmers used to study cooperative motion underwater.[40]

**Results and discussion**

*Robot design*

The BackBot compartments are depicted in **Figure 2**. It is constructed of a main 3D-printed hollowed chamber (main body), with an onboard electrode at the top and a 3D-printed cellular canopy on top of the electrode (Figure 2a).

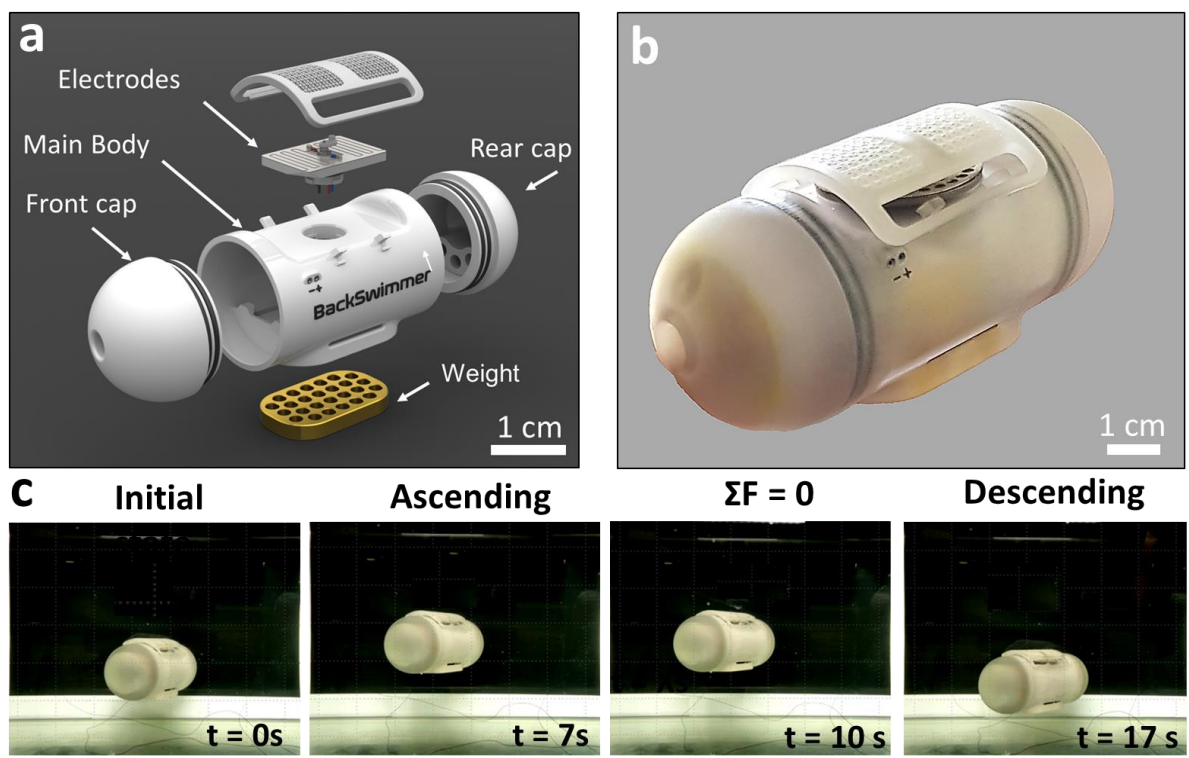

**Figure 2.** a) Backswimmer-inspired robotic design: main body, an onboard electrode on top, a linearly movable cellular structured canopy, and a brass "trimming" device for adjusting weight and center of mass. b) The 3D printed BackBot. c) Four stages in the auto-regulation of the BackBot buoyancy: i) initial stationary state, ii) bubble nucleation and growth through catalytic activity, resulting in motion upward, iii) controlled release of bubbles in order to achieve force balance and vi) further release of bubbles reverse the forces, resulting in descending of the robot.

A brass plate with twenty-six threaded holes was positioned at the bottom outer part of the robot's hull (see also **Figure S1** in the supporting information). A rubber O-ring is fitted inside a groove on the bottom side of the platform, to prevent water leakage to the robot interior volume. The buoyancy device assembly was designed to be easily dismounted from the robot. The bottom weight allows to stabilize the robot by lowering the center of gravity underneath the center of volume. Fine adjustments are made to reach a desired density by adding or removing setscrews from the brass plate, namely "trimming". In addition, the set of weights is used to compensate for deviations between the center of gravity and the center of volume line of action, and align the robot to a stable spatial orientation in the bulk water. An optical figure of the BackBot is shown in Figure 2b. The overall weight of the robot is 112 g and it is mostly made of 3D printed polymeric parts. Its average density, $\rho_{robot}$= 1.0025 g·cm$^{-3}$, was designed to be slightly larger than the density of the surrounding water in order for the robot to be negatively buoyant in its non-active state. Figure 2c shows the principle of operation of the BackBot and the buoyancy auto-regulation. Initially, the BackBot is situated at the bottom of a water tank. Upon applying voltage, bubbles nucleate on the electrodes. Some bubbles remain anchored to the electrode, and some leave the surface and float until they reach the canopy. There, they become entrapped and are stabilized at the solid-water interface. Consequently, the BackBot becomes positively buoyant and starts ascending. In order to balance the positive buoyancy forces, the linear oscillation mechanism starts operating, causing bubbles to release from the surface of the canopy. This step is iterative, and eventually leads to neutral buoyancy, when the overall buoyancy force equals the gravitational force derived from the weight of the BackBot.

*Microbubbles nucleation, growth and entrapment through electrolysis*

The buoyancy auto-regulation device is depicted in **Figure 3**. At the top of the device a 3D-printed canopy, made of a curved lattice structure, is made to entrap bubbles. A crank arm, mounted on a DC motor shaft, converts the rotational movement generated by the motor to linear movement of the canopy (Figure 3a). Two electrodes, made of 0.2 mm thick stainless steel in comb configuration, produce gas bubbles through electrolysis when voltage is applied to it. A 3D-printed platform was designed to accommodate the DC motor and electrodes. A cross section of the canopy reveals the lattice structure used for capturing bubbles (Figure 3a, i). The canopy was designed as a curved frame, supporting an inner cellular structure, constructed of a repeating unit cell. Each unit cell is shaped as an extruded ring sector, which follows the canopy curvature. The extruded dimensions corresponds to 2.5 mm, sector angle of 6 °, and thickness of 2.5 mm.

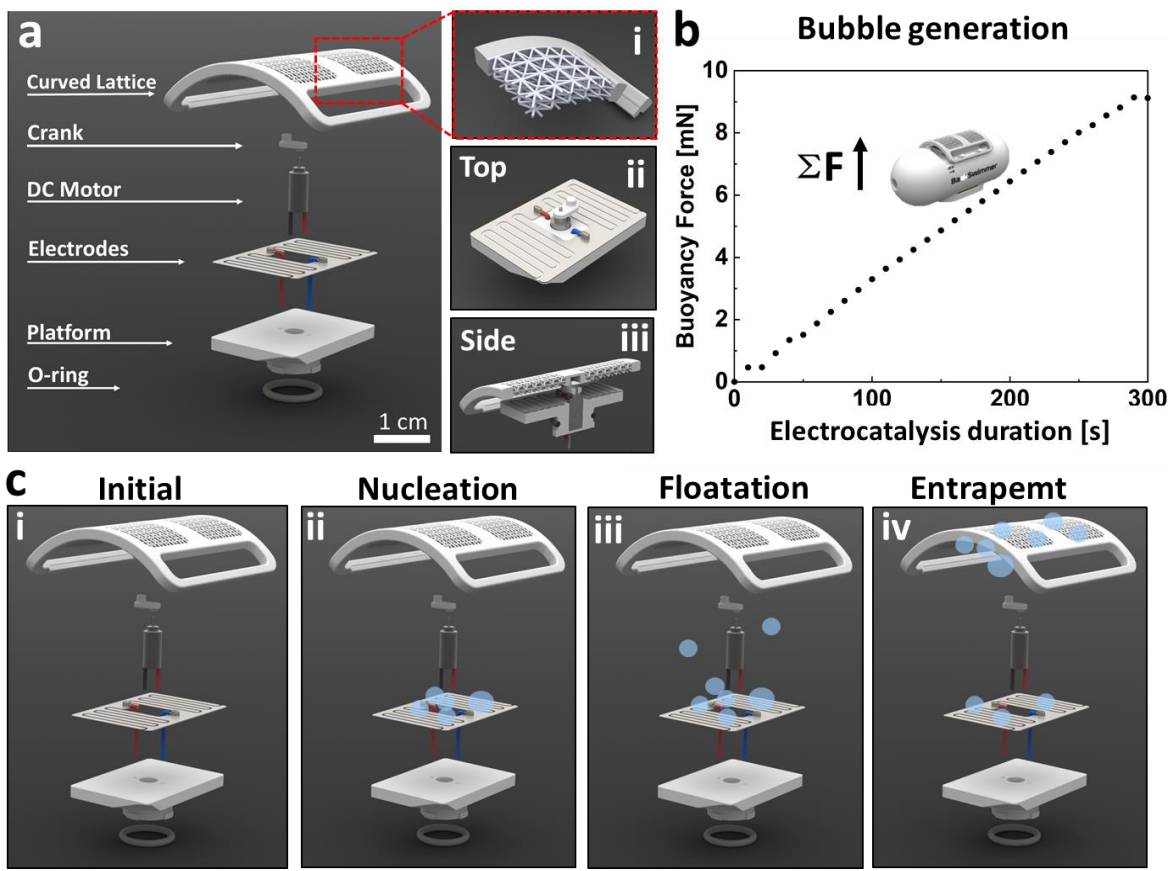

**Figure 3**. a) Controlled bubble nucleation and growth unit, and a cellular 3D printed canopy (i) for bubble entrapment. Two interlocking stainless steel electrodes comprise two electrolysis sites for microbubbles nucleation (ii). b) Buoyancy force depending on the electrolysis duration. c) Bubbles reaching a critical size detach from the surface and entrap in the cellular 3D printed canopy above.

Three different unit cells were designed, fabricated and tested in order to achieve optimal capturing of the gas bubbles as they pass through the lattice voids and get attached to the surface of the round rods. A hybrid body-centered cubic (BCC) and face-centered cubic (FCC) unit cells of 2.5 mm edge length, showed a good capturing ability of the gas bubbles (**Figures S2** and **S3** in the supporting information). Specifically, only a fraction of microbubbles managed to pass through the bars without being trapped.

A top view of the buoyancy device showing the crank arm, DC motor and electrodes is depicted in Figure 3a (ii). The electrodes comprise two interlocking combs with 0.5 mm spacing between their scales. This geometry maximizes the surface area for bubble nucleation, and increases the electrostatic fields between the combs. This configuration yields small gas bubbles of up to 1 mm in diameter. Such bubbles detach spontaneously from the electrodes and float upwards to the canopy. The importance of bubble detachment from the electrodes lies in the difficulty to release them from this surface in a controlled fashion. A cross section of the assembled configuration of the buoyancy regulation device is depicted in Figure 3a (iii). While the canopy is designed to linearly vibrate and regulate bubble release, the electrodes cannot be freed from anchored bubbles. This results in two counterproductive outcomes. First, anchored bubbles reduce the electrolysis efficiency, and consequently, the ability to generate maximal buoyancy forces. Second, anchored bubbles alter the density balance of the device and may interfere with the control loop mechanism regulating the buoyancy as will be discussed in the next parts. Figure 3b shows a representative curve of the total forces acting on the BackBot, depending on the duration of applied voltage. The overall force increases linearly with time, in a rate of 0.031 mN·s$^{-1}$ (see supporting information, **Figure S4**). Figure 3c illustrates the operation of the device. Microbubbles nucleate and grow on the surface of the electrodes, detach and float upwards, and become entrapped in the canopy situated above.

*Bubble release through linear vibrations*

The bubble controlled release mechanism is based on linear oscillations of the canopy (**Figure 4**). The gas bubbles, anchored to the canopy, are subjected to lateral forces, exerted by oscillations at a frequency of 10 Hz (Figure 4a, inset). This frequency is high enough to induce detachment of bubbles, but small enough to be compatible with the available power, provided by the battery. The power consumption of the motor corresponds to 1.875 W.

This controlled release mechanism decouples the oscillations of the canopy from the main body of the BackBot, and inhibits destabilization of the device. Figure 4b depicts the conversion mechanism from rotation to linear motion. A crank arm is fitted on a DC motor shaft, and a pin on the crank arm is allowed to move inside an embedded slot in the canopy part. This pin-slot configuration converts the rotational motion of the motor shaft to linear motion of the canopy (Figure 4c). As the canopy oscillates around the center position, gas bubbles detach from the structure and are released to the bulk water.

The capillary length of water in air is given by $l_c = \sqrt{\gamma/\rho g}$, with *ρ*- the water density and *g*- gravitational acceleration. For droplets in air, this corresponds to roughly 2.7 mm. We expect that the capillary length calculated for gas bubbles underwater will be similar. Meaning, bubbles below this size will remain attached to the canopy, while larger bubbles will detach spontaneously due to buoyancy forces overcoming the capillary adhesion forces. Indeed, microbubbles, up to two millimeters in diameter, show significant pinning on the lower bars of the canopy.

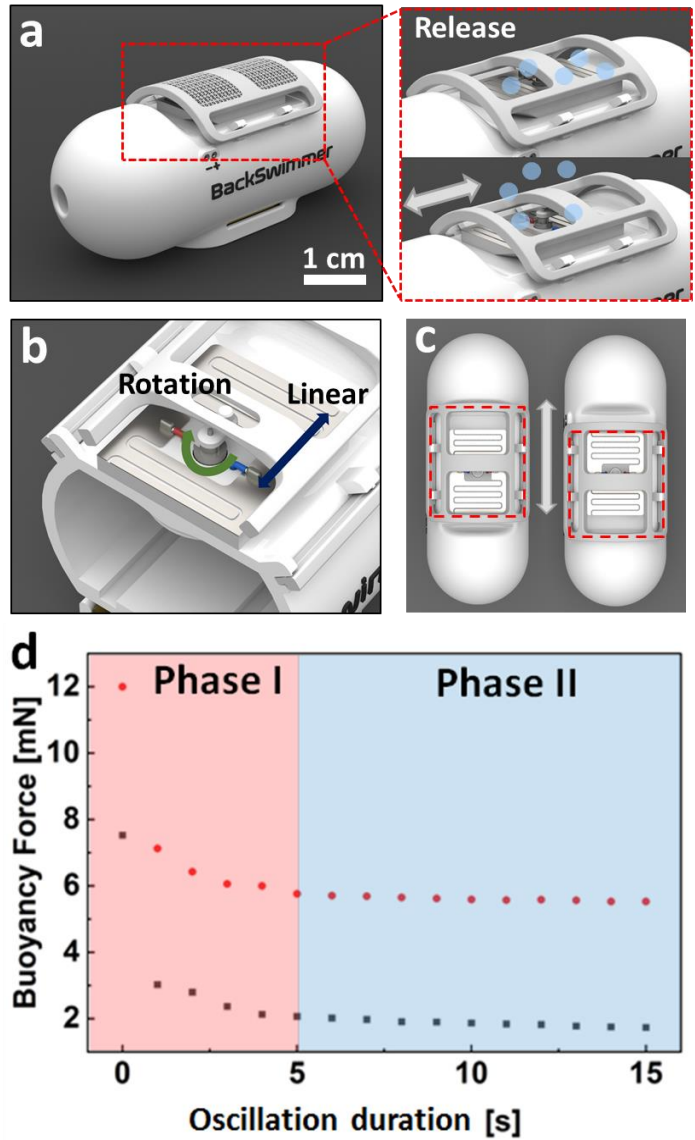

**Figure 4**. Bubble release mechanism through linear mechanical vibrations. a) The cellular 3D printed canopy linearly oscillates to release surface anchored bubbles (inset). b) A pin-slot configuration converts the rotational motion of the motor shaft to linear motion of the canopy. c) Top view of the canopy, oscillating in frequency of 10 Hz, over a span of 6 mm. d) Buoyancy force decrease, depending on the oscillation duration.

Figure 4d shows the characteristic force curves, depending on oscillation duration. Bubbles detachment is divided into two phases. In the first phase (red), up to approximately 5 seconds of oscillations, an abrupt decrease in buoyancy force occurs due to coalescence and release of bigger bubbles from the canopy. In the second phase (blue), there is a minor decrease in the buoyancy force due to minimal release of small bubbles. The two sets of data (red, black) show two representative experiments with different initial conditions, in terms of amount of gas entrapped in the canopy. In both cases, the duration of the first phase and characteristic decrease in force are comparable. The overall force decreases with oscillation time, with maximal rates of 0.34 ± 0.04 mN·s$^{-1}$ in phase I, when the canopy is fully saturated with bubbles, and 0.027 ± 0.006 mN·s$^{-1}$ in phase II. We, therefore, adjusted the operation of the Backbot to work with partial occupancy of the canopy with bubbles (phase II).

The lateral adhesion of a bubble to a surface is given by **Equation 1**:[41,42]

$$F_{Lateral} = \gamma L_s (\cos\theta_{Rec} - \cos\theta_{Adv}) \tag{1}$$

$F_{Lateral}$ is the lateral adhesion force, acting at the contact of the bubble with a solid surface, $\gamma$ is the interfacial tension of the liquid, $L_s$ is the contact width and $\theta_{Rec}$ and $\theta_{Adv}$ are the receding and advancing contact angles, respectively. The advancing and receding contact angles of the 3D-printed material the canopy is made of correspond to $\theta_{Adv} = 113 \pm 6°$ and $\theta_{Rec} = 71 \pm 12°$, respectively. This results in adhesion forces of approximately few µN to few dozens µN for bubbles ranging between 100 µm - 1 mm in diameter, respectively. While the adhesion forces scale linearly with the bubble radius, the mechanical forces, due to oscillations, scale with the bubbles volume. Therefore, the estimated forces acting on the surface anchored bubbles due to the mechanical oscillations correspond to roughly ~10$^{-3}$ µN for bubbles with 100 µm diameter but up to µN scale for bubbles with 1 mm in diameter. For bubbles larger than 1 mm in diameters, buoyancy forces become more dominant, in addition to the lateral adhesion forces. In practice, this means bubbles of up to few hundred microns in diameter cannot be mechanically released from the canopy, while only bubbles of roughly 1 mm in diameter and above release due to the oscillations. This agrees very well with the description of phase I and phase II (Figure 4d). In phase I, the larger bubbles detach, causing quick reduction of buoyancy forces, followed by phase II, in which smaller bubble remain attached to the canopy, resulting in only small residual decrease in the buoyancy force.

The electrolysis power consumption corresponds to 0.75 W. In the ascending phase, energy consumption from initial bubble production until floatation corresponds to roughly 23 J. Further changes in depth correspond to energy consumption of 18 J in average per transition between states, within a range of approximately 30 cm tested in these experiments. Overall, the Backbot ascends and descends while consuming power in the range of dozens of Joules, compared to energy consumption of hundreds and thousands of Joules in other devices[43–45].

Note the difference between the time scales for buoyancy force generation of couple hundreds seconds (Figure 3c), as opposed to bubble release of few seconds (Figure 4d). These correspond to the maximal buoyancy forces that can be generated, on one hand, of approximately 10 mN, but only to incomplete bubble release, resulting in force decrease of roughly 6 mN, on the other hand. In order to match the time scales for nucleation and release, we defined a force working range of approximately 1 mN, depending on the trimming. With this working range, characteristic times for generating positive buoyancy forces correspond to a few dozen seconds and few seconds for transitioning from positive to negative buoyancy. In practice, during the first cycle of bubble generation, the canopy becomes saturated with bubbles so that the force working range reduces. Consequently, the time scale for transitioning between positive and negative buoyancy states reduces to few seconds (Figure S4 in the supporting information).

*Buoyancy auto-regulation*

Finally, we introduce a closed control loop using a pressure sensor and a Proportional Integral Derivative (PID) controller (**Figure 5**).

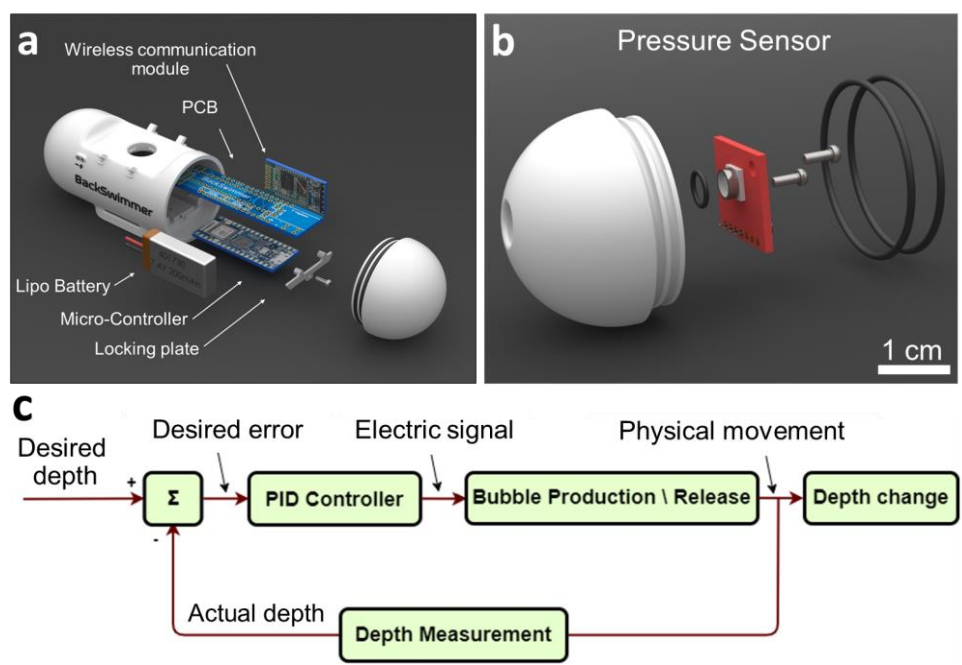

**Figure 5.** An exploded view of the main electronic components. a) Custom-made PCB board, containing the voltage regulator, H-bridge integrated circuits, resistors and miniature electrical connectors, Bluetooth wireless communication module, Lipo battery and an Arduino Nano microcontroller. b) The front cap, showing the pressure sensor, two securing screws and rubber O-rings for sealing. c) A block diagram showing the control scheme of the system.

The main electronic components include a custom-made printed circuit board (PCB) with the necessary electronic components such as the voltage regulator, H-bridge integrated circuits (ICs), resistors and miniature electrical connectors (Figure 5a). A wireless communication module allows communication between the BackBot and a remote station. An external voltage supply port allows bypassing the battery when long experimentation time is needed. Figure 5b shows an exploded view diagram of the pressure sensor, the front cap, and the two securing screws and rubber O-rings for sealing the system. Figure 5c shows a sketch of the automation process governing the buoyancy regulation. The closed loop control algorithm, operating at a rate of 10 Hz, receives an input parameter from the operator (desired depth) and depth measurement from the pressure sensor (actual depth). The difference between the two parameters is the "depth error", fed into PID control algorithm. The output of the algorithm is a value ranging between -255 and +255, corresponding to the electrical signal intensity. This is sent to the electrodes and motor driver in order to produce or release bubbles from the buoyancy device, accordingly. During this process, the change in depth is achieved by physical movement of the robot along the z-axis (see also **Figure S5** in the supporting information).

We then characterize the BackBot ascending and descending rates over time (**Figure 6**). The zero position corresponds to the water-air interface. Ascending rates correspond to $11.3 \pm 2.7$ mm·s$^{-1}$ (Figure 6a). This rate depends on the duration of applied voltage and is tunable. In the opposite direction, when the BackBot descends, rates correspond to $31.5 \pm 22.4$ mm·s$^{-1}$ (Figure 6b). Variations in the descending rate are the result of residual entrapped gas bubbles in different locations of the robot, including the canopy, main hull and external areas, and could not be entirely avoided. Figure 6c shows controlled ascending and descending of the BackBot until it reaches neutral floatation, in an autonomous way, through the controlled PID loop. The vertical range for auto-correction of the BackBot position is roughly 2 cm. Consequently, it remains stable in a fixed vertical position, limited by the diffusion rates of the gas from the surface anchored microbubbles into the bulk water, and up to few hours.[46]

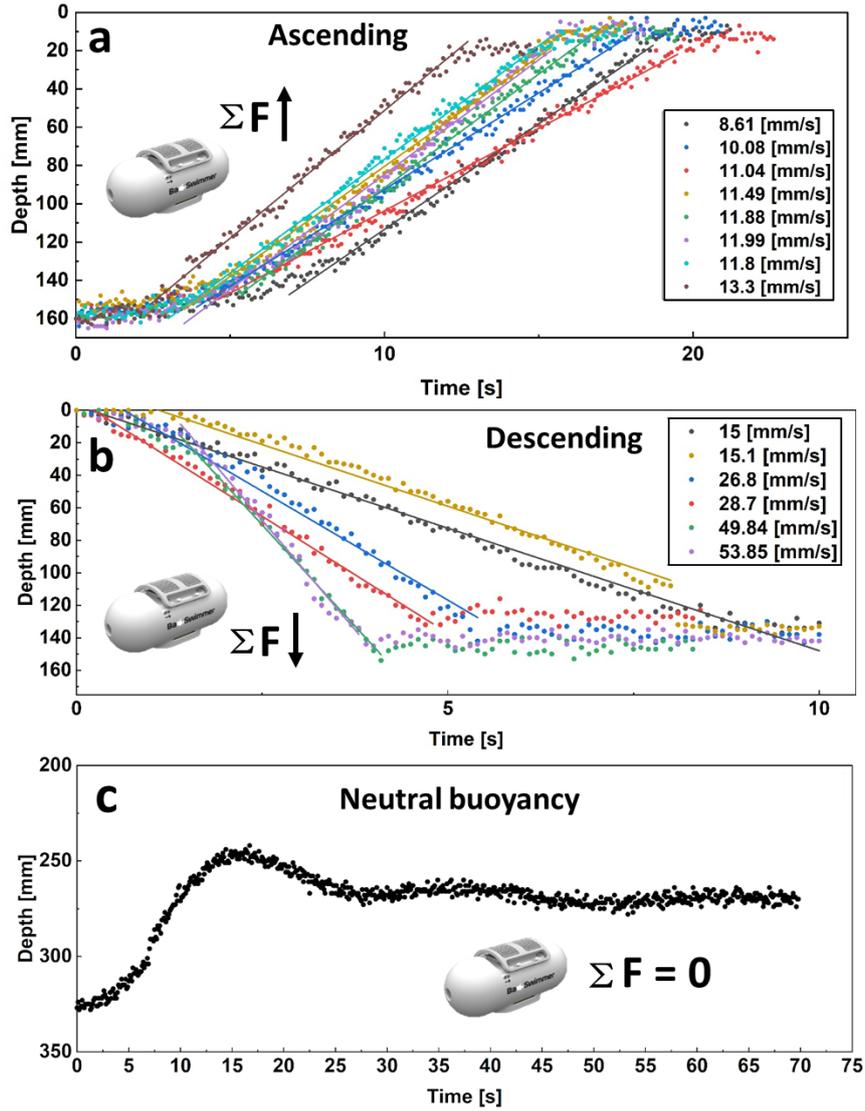

**Figure 6.** Buoyancy regulation of the BackBot during a) ascending, b) descending and c) reaching neutral floatation.

**Figure 7** shows a variety of motion patterns in the vertical direction, generated by manual control (Figure 7 a,b) and autonomous control (Figure 7 c,d). In the manual mode, two control signals are chosen by the operator: one for the electrodes and one for the motor generating the mechanical vibrations. These are sent from the remote station to the robot wirelessly. Average ascending and descending velocities correspond to $17.7 \pm 0.7$ mm·s$^{-1}$ and $23.5 \pm 2.5$ mm·s$^{-1}$ for the first ($t_1+t_2$) and second ($t_3+t_4$) cycles, respectively (Figure 7a). In Figure 7b, the potentiometer controlling the electrodes (POT-E) and the DC motor (POT-M) were controlled more precisely, leading to a more moderate movement of the robot. A first ascend ($t_1$) started at a depth of 150 mm up to a depth of 75 mm, at an average velocity of 6.8 mm·s$^{-1}$. Then, the BackBot stabilized ($t_2$). A rapid movement of the canopy made the robot sink at an average velocity of 48 mm·s$^{-1}$ ($t_3$) until it reached the bottom of the water tank ($t_4$). Bubble production was renewed (t$_4$), and the robot ascended to a depth of 75 mm at an average velocity of 14.5 mm·s$^{-1}$ (t$_5$), for 4 s (t$_6$). Finally, the BackBot climbed to a depth of 49 mm and maintained its position at this depth (t > t$_6$).

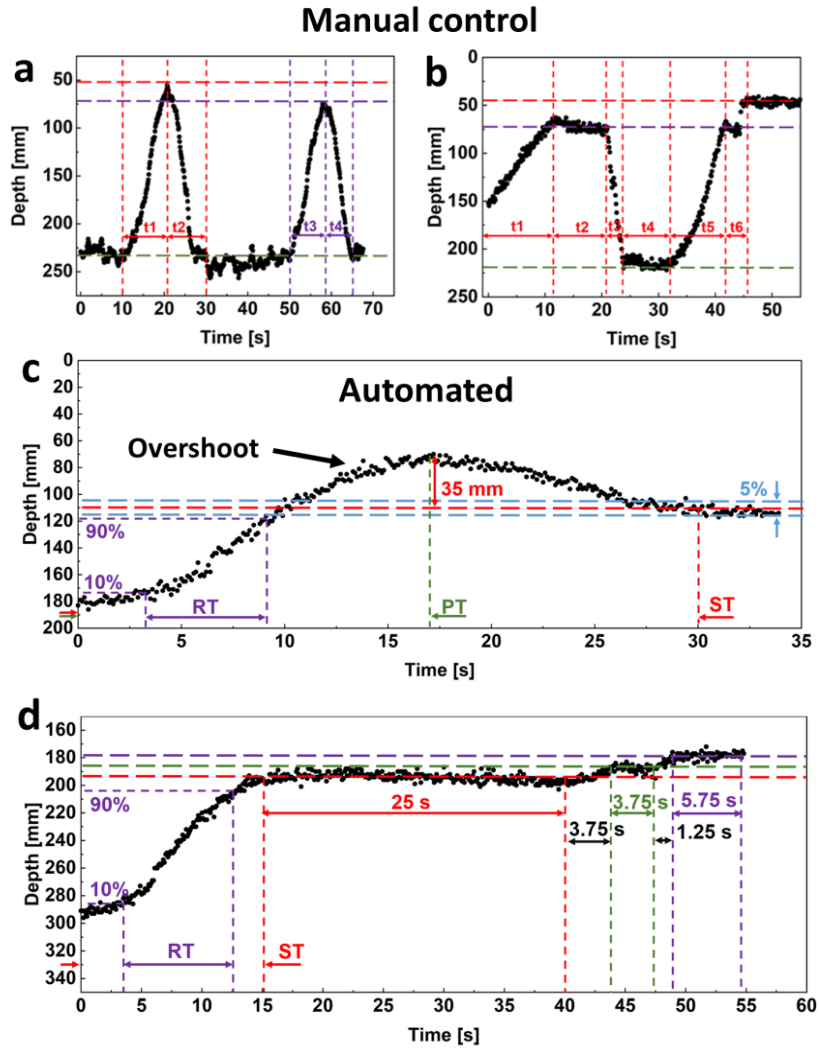

**Figure 7.** BackBot performance in manual and automated modes. Depth profiles that include a) rapid ascending and descending and b) ascending, descending with delay time in between. c) Depth profiles of automated buoyancy regulation with overshooting and d) precise accomplishment of the target depth. Reducing overshoot is achieved by tuning the PID closed loop.

In the automated mode, using the closed loop PID control algorithm, a target depth was chosen by the operator and sent to the BackBot. We tune the gain factors of each term in the PID control algorithm: proportional term gain- $K_p$, integral term gain- $K_i$, and derivative term gain- $K_d$, in order to improve the motion control in terms of overshoot, oscillations and time until the target depth is reached. At first, values of $K_p = 10$, $K_i = 0$, $K_d = 0$ were used. This resulted in an overshoot of 35 mm and reaching the target depth (within 5 % error) after 30 s (Figure 7c). Additional tuning of the proportional and integral parameters improved the performance.

Introducing the derivative parameter, $K_d$, minimized the overshoot and lowered the setting time. The values were now set to $K_p = 2.5$, $K_i = 0.9$, $K_d = 0.1$ (Figure 7d). Consequently, the rise time reduced to 9.1 s with no overshoot. The setting time required for the BackBot to reach the target depth, within an error of 5 %, reduced to 15 s. Furthermore, after the target depth was reached, new target depth was sent to the robot in order to test how the system responds to changes. After hovering at the first target depth (Figure 7d, horizontal dashed red line) for 25 s, the robot ascended to the new target depth (horizontal purple dashed line). The intermediate green dashed line indicates a disturbance that the buoyancy device was able to overcome, when small bubbles managed to escape the canopy (t = 43 s - 46.75 s). The controller was able to compensate for the loss and reach its target depth. The ascending and descending rates depend on the initial conditions, such as the preload and trimming, resulting in different ratios of mass to volume of the Backbot.

*Summary and conclusions*

In this work, we presented an automated mechanism for controlling buoyancy of an underwater device. This mechanism does not rely on inflation and deflation of an air tank, does not involve the use of pistons but rather controlled generation and release of surface anchored bubbles. Bubbles are generated through electrolysis of the surrounding water, directly on on-board electrodes. They are then transferred into a canopy, made of a 3D cellular structure, which can control their release into the bulk water through mechanical vibrations. We introduce a closed loop control algorithm in order to induce buoyancy self-regulation. We show that by tuning specific parameters, the robot reaches a target depth within few seconds, over a depth difference of few dozen centimeters. In outlook of this work, this mechanism will be integrated in an underwater robotic swimmer that will be able to propel and propagate underwater. In a broader perspective, this platform will be used for developing a group of self-propelled underwater swimmers and for investigating the interactions between them, including fulfilling mutual tasks, collaboration and reciprocal processes.

**Materials and Methods**

*3D Printing*

The robot frame and buoyancy device components were fabricated using additive manufacturing technology (MJP-2500 plus, 3D Systems, voxel size of approximately 25 μm) using VisiJet® M2R-CL photopolymer. All parts were post-processed in accordance with the manufacturer guidelines and include immersing the parts in mineral oil bath and soap water 60 °C. All parts were coated with silicone grease (Dow Corning® high vacuum silicone grease) to ensure good sealing.

*Prototype fabrication and assembly*

The 3D models were designed and evaluated using Solidworks CAD software. CNC Laser cutting technology (Fiber Laser) was used to cut the electrodes (stainless steel 304, Scope Metal Group Ltd.). The electrodes were glued with double-sided adhesive or instant glue. The laser cutter was also used to cut sheets of Poly-(methylmethacrylate) (PMMA) for various experimental setup jigs. CNC milling and turning processes was used to manufacture metal parts with complex geometry. PCB manufacturing process was used to produce the custom made electrical circuit board needed for the project (JLCPCB Ltd.).

*Hardware and electrical components*

Two cells, 7.4 V, 200 mAh, rechargeable Lipo battery provided the necessary energy for the robot. The wireless communication module establishes full-duplex communication channel with the microcontroller via UART protocol, and was paired with another module located on the PC side via Bluetooth protocol. The depth reading was done with a pressure sensor (TE MS5803-14BA) mounted on the robot body and was interfacing directly with the surrounding water. The pressure sensor establishes full-duplex communication channel with the microcontroller via I2C protocol. Arduino Nano 33 BLE Sense Microcontroller was used. Power consumption when only controller and communication modules operate correspond to 0.26 W.

*Force measurements underwater*

The bubble nucleation and buoyancy experiments were recorded with a digital camera (OLYMPUS A1) and were analyzed manually. The current and voltage were measured with a digital multi meter (Tektronix DMM4050). The overall force acting on the submerged BackBot underwater was measured using a digital balance scale (Fischer Scientific), see supporting information, **Figure S6**. The measurements were performed at specific times intervals, depending on the time scale of interest.

**Acknowledgements**

This research was supported by the Ministry of Science & Technology ,Israel, grant number 3-17384, and by the Zimin Institute for Engineering Solutions Advancing Better Lives.

# Supporting Information

**Backswimmer Inspired Miniature Robot with Buoyancy Auto-Regulation through Controlled Nucleation and Release of Microbubbles**

Dror Kobo, Bat-El Pinchasik*

A brass plate, 26 g in weight, with 26 m5-threaded holes (Figure S1) was positioned outside the device main body, at the bottom side of the robot hull, to allow for quick access during experiments and to give stability to the robot by lowering the center of gravity beneath the center of volume. Fine adjustment were made to reach a desired density by adding or subtracting screws to the brass plate, namely "trimming". Another function of the set screw weights is to compensate for deviations of the center of gravity from the center of volume in the horizontal plane and align the robot spatial orientation.

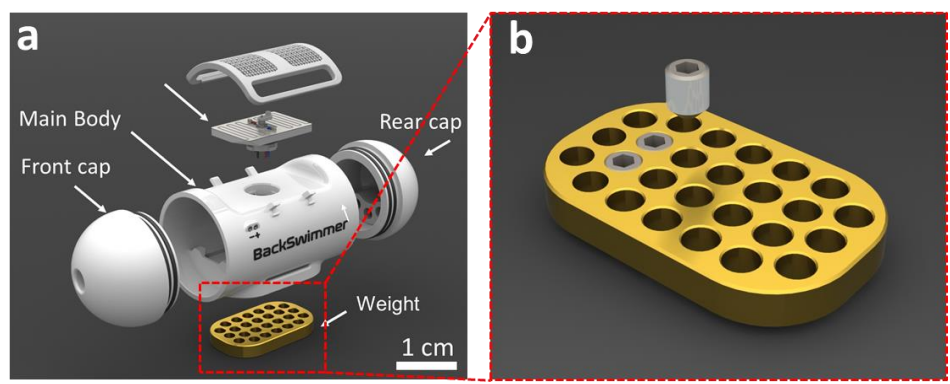

**Figure S1.** Density adjustment unit with possible modulation of center mass. a) The location of the brass plate, at the bottom of the BackBot. b) Higher magnification of the brass plate sketch and screws used to adjust the device center of mass.

Cellular canopy structures tested for bubble entrapment and controlled release (**Figure S2**). Round bars (Figure S2 a), oval holes (Figure S2 b), 0.3 mm rod thickness BCC lattice design (Figure S2 c), and 0.3 mm rod thickness BCC/FCC lattice (Figure S2 d). The latter showed the best performance in terms of capturing capability of microbubbles, their stabilization and controlled release during linear oscillations.

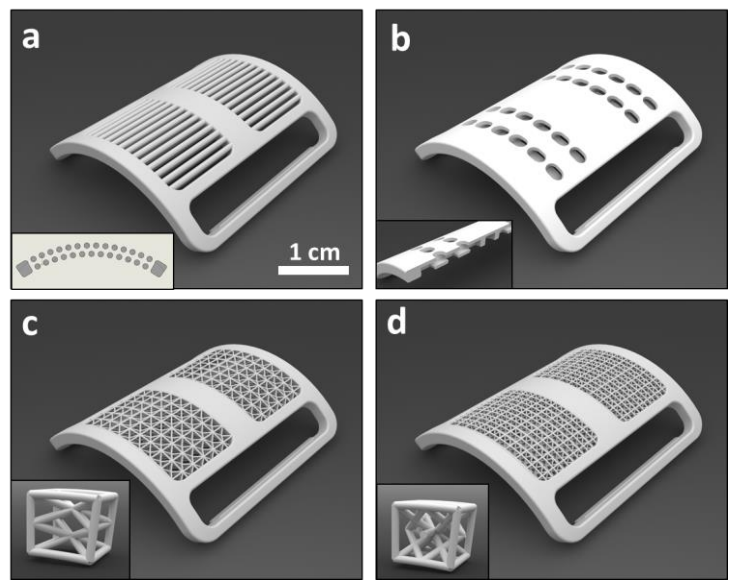

**Figure S2.** Designs of bubble capturing 3D printed canopies. a) Round bars, b) oval holes, c) 0.3 mm rod thickness in BCC lattice, d) 0.3 mm rod thickness in a BCC + FCC lattice.

For the cellular structure of the canopy, three designs were tested: curved body centered unit cell with rod diameter of 0.3 mm, 0.4 mm and a combination of body centered and face centered unit cells with rod diameter of 0.3 mm (**Figure S3**). The latter yielded more sites for bubbles to attach and grow, and better bubble detachment than the other geometries. Although the latter design has smaller surface area than the other structures, it presented better results in terms of anchoring gas bubbles. This is due to higher number of anchoring sites that enabled microbubbles to entrap more easily within the structure, but also to percolate upwards during vibrations.

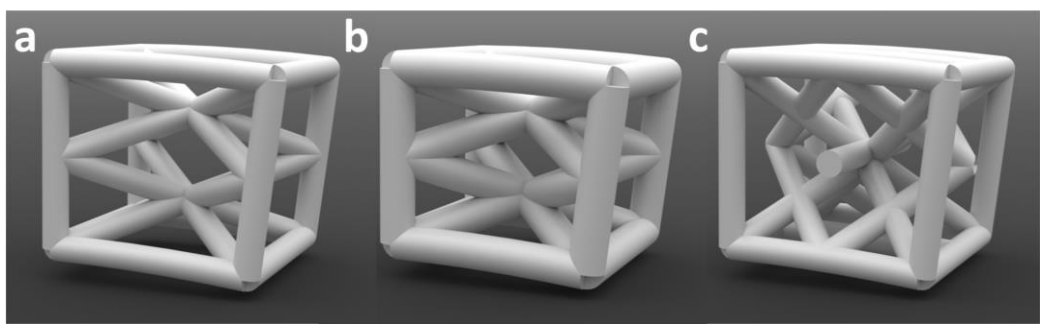

**Figure S3.** Curved unit cells used for the canopy cellular design. a) 0.3 mm rod thickness with double BCC unit cells on top of each other, with total surface area of 8274.3 mm$^2$. b) 0.4 mm rod thickness with double BCC unit cells on top of each other, with total surface area of 9493.5 mm$^2$. c) 0.3 mm rod thickness with combined BCC and FCC unit cells, with total surface area of 9234.7 mm$^2$.

Bubble production and release, and the consequent buoyancy force generation and reduction, respectively (**Figure S4**). The buoyancy force generation (black) is linear with time and equals 0.031 mN·s$^{-1}$. Upon bubbles release, an initial sharp drop in buoyancy force is followed by a moderate decrease in force, in a steady state. In this phase, the rate corresponds to -0.026 mN·s$^{-1}$. Bubbles nucleation (black) results in linear increase of buoyancy forces in a rate of 0.031 mN·s$^{-1}$.

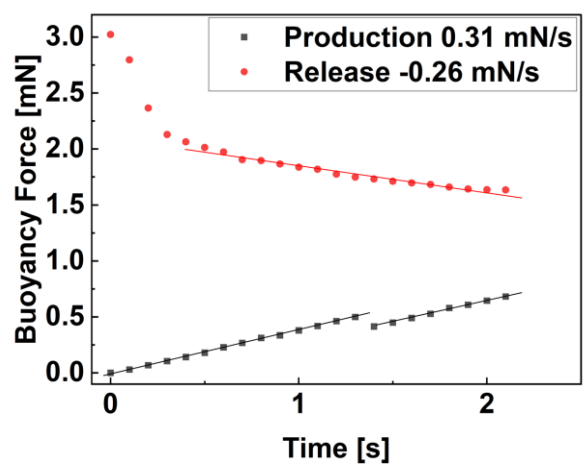

**Figure S4.** Buoyancy force generation (black) vs. elimination (red).

The PCB is mechanically and electronically connected to the microcontroller unit (MCU) through metal pins connectors. Additional parts include HC-05 Bluetooth wireless communication module, two-cells 7.4 V, 200 mAh Lipo battery, Arduino Nano 33 BLE Sense microcontroller, a locking plate and a screw to align and secure the electronics to the robot hull. A voltage regulator, mounted on the PCB, reduces the voltage to 5 V, in order to fit the input voltage level of the microcontroller and other peripheral components.

A Proportional Integral Derivative (PID) closed loop control algorithm, was used in order to induce auto-regulation of buoyancy (**Figure S5**). A target depth is determined (set point), the actual depth of the robot is measured by the pressure sensor (process variable), and the difference between the set point and the process variable is defined as the error. The error is then fed into the PID controller. The proportional controller output is a control signal, proportional to the error value at a given time. The integral controller outputs a signal, proportional to the accumulated error of the system. The derivative controller outputs a control signal, which is the estimate of the future trend based on the current rate of error. The three controllers' signals are then summed up and sent to the appropriate motor or electrodes' drivers as a Pulse Width Modulation (PWM) signal. The PID controller is running continuously at fixed time intervals.

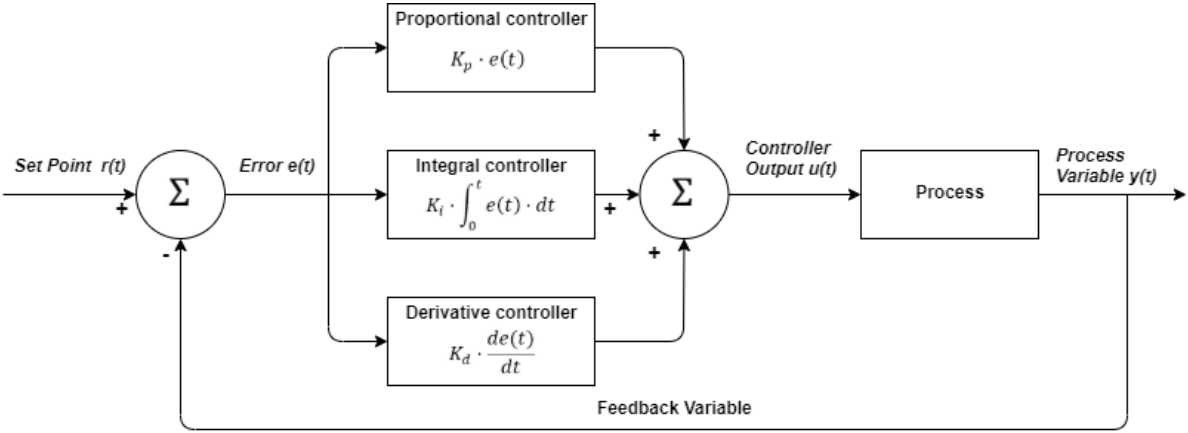

**Figure S5**. PID (Proportional, Integral and Derivative) closed loop control algorithm scheme. In this control algorithm, a set point "r(t)" is predetermined by the user. The Error "e(t)" is the difference between the set point and the feedback variable "y(t)", which is measured by a sensor. The error is fed to the three separate controllers and calculated. The output is than summed to a single output "u(t)", converted to an electrical signal and sent for the process to be executed. The process is iterative at predefined execution frequency.

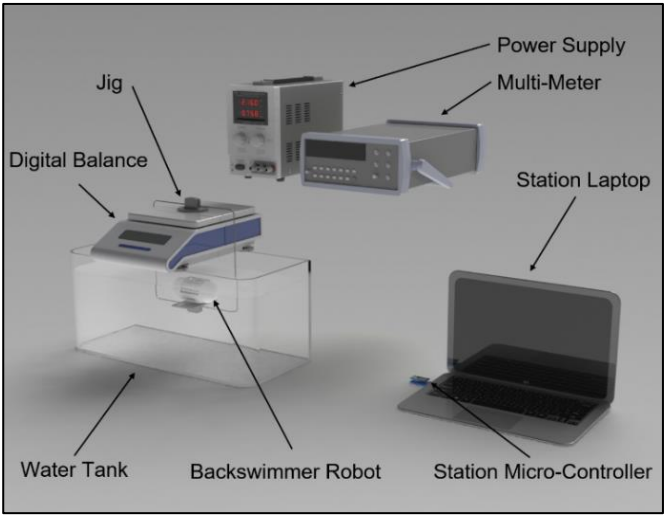

**Figure S6.** Experimental setup of the buoyancy force measurements. The depth control experiments were done without the digital scale and jig. (The laptop, digital balance, multi-meter and power supply CAD models were taken from grabCAD Inc).